\documentclass[12pt,a4paper,onecolumn]{article}
\usepackage[utf8]{inputenc}
\usepackage[english]{babel}

\usepackage{times}
\usepackage{epsfig}
\usepackage{graphicx}
\usepackage{amsmath}
\usepackage{amssymb}
\usepackage{rotating}
\usepackage{multirow}
\usepackage{subfig}
\usepackage[T1]{fontenc}
\usepackage{times,multicol,graphicx}
\usepackage{xcoffins}
\usepackage{times,color}
\usepackage[left=2cm,right=2cm,top=2cm,bottom=2cm]{geometry}
\author{Valerio Mura, Giulia Orr\`{u}, Roberto Casula, Alessandra Sibiriu, \\Giulia Loi,  Pierluigi Tuveri,
Luca Ghiani, and Gian Luca Marcialis\\
University of Cagliari - Department of Electrical and Electronic Engineering - Italy\\
{\tt\small \{valerio.mura, giulia.orru, pierluigi.tuveri,} \\
{\tt\small luca.ghiani, marcialis\}@diee.unica.it}}
\title{LivDet 2017 Fingerprint Liveness Detection Competition 2017}

\begin{document}
\maketitle
\begin{abstract}
Fingerprint Presentation Attack Detection (FPAD) deals with distinguishing images coming from artificial replicas of the fingerprint characteristic, made up of materials like silicone, gelatine or latex, and images coming from alive fingerprints. Images are captured by modern scanners, typically relying on solid-state or optical technologies. Since from 2009, the Fingerprint Liveness Detection Competition (LivDet) aims to assess the performance of the state-of-the-art algorithms according to a rigorous experimental protocol and, at the same time, a simple overview of the basic achievements. The competition is open to all academics research centers and all companies that work in this field. The positive, increasing trend of the participants number, which supports the success of this initiative, is confirmed even this year: \textbf{17} algorithms were submitted to the competition, with a larger involvement of companies and academies. This means that the topic is relevant for both sides, and points out that a lot of work must be done in terms of fundamental and applied research.
\end{abstract}

\section{Introduction}
    \vspace{-5pt}
In the last years, fingerprint recognition applications reached a large portion of non-expert users thanks to the grow up of the mobile devices market. Solid-state-based micro-scanners have been added to such devices, thus encouraging the dissemination of fingerprint verification checks and the confidence of the users on this biometric characteristic. However, artificial fingerprint replicas, also called spoofs, or fake fingerprints, make possible to circumvent fingerprint-based personal recognition systems. This problem involves both border control applications and personal identity verification systems in the broad users market, with serious concerns when these systems are conceived as part of critical infrastructures.
\\A Fingerprint Presentation Attack Detector or Fingerprint Liveness Detector (FLD) is a machine learning-based system able to prevent direct attacks to the scanner, by discriminating images captured from live fingers and those coming from fake ones. FPAD can be software based or hardware based. If additional hardware may assure a better liveness detection rate, by measuring the blood pressure or the heartbeat, it may be difficult to update and improve the system, if attacks are designed to overcome such additional sensors. On the other hand, software-based approaches, which predict the liveness degree of a certain input given the captured image, are easier to``patch'' than hardware-based ones. This is crucial when considering that the liveness detection is an arms-race problem where the attackers are expected to gradually refine and improve their fake fingerprint fabrication ability and techniques. As a plethora of papers showed, dealing with the liveness detection problem by image processing and machine learning methods, that is, software-based methods, is not trivial at all \cite{ReviewLivDet,Marasco:2014:SAS:2658850.2617756}. 
Moreover, there is no general agreements about the experimental protocol and tests to be carried out for a correct assessment of the FPADs performance. From this point of view, a great help has been provided by the International Fingerprint Liveness Detection Competition editions from 2009 to this last one, whose this report summarizes the basic results \cite{ReviewLivDet}. The greatest contribution of each competition edition is to provide up to four data sets of live and fake fingerprint images in order to support the research community efforts in improving the FPADs' performance. Moreover, an established experimental protocol is adopted to allow the fair comparison of the reported performance. Finally, the most modern hardware technology, the finest fake fingerprint fabrication techniques and novel materials are possibly added every edition. Participation of academies and companies is always welcome and encouraged. We are very proud to see that the fundamental and applied research communities are growing up even thanks to the big efforts necessary to organize a competition like this, where expert people is involved in fabricating thousand of spoofs, made publicly available by images captured with different capture devices. 

The goal of this report is to describe the LivDet 2017 competition characteristics and to summarize the results achieved by the  participants. Section \ref{sec:Overview} gives an overview of what we have done in this competition edition. The evaluation protocols and participants are reported in Section \ref{sec:Protocol}.  Section \ref{sec:Result} discusses the competition results and Section \ref{sec:Conclusion} concludes the paper.

\section{Fingerprint Presentation Attack Detection}
    \vspace{-5pt}
\label{sec:Overview}
The first paper on fingerprint presentation attacks dates back to 1998, when D. Willis and M. Lee experienced that four scanners out of six were ``vulnerable'' to spoofs placed on the acquisition surface \cite{Scanners}. 
Later, Van der Putte \cite{vanderPutte2000} and Matsumoto \cite{MatsumotoMYH02} independently confirmed this issue, by extending the investigation to other capture devices. Both research groups focused on the vulnerabilities of a fingerprint recognition system caused by falsification of fingerprints.
As well as the problem appeared as evident, countermeasures were soon proposed. The first hardware-based solution and the first software-base solution were published \cite{kalloPatent,Schuckers02spoof}. From those years, fingerprint presentation attacks detection has become one of the hottest topics in the last decades, especially from the software-based approaches viewpoint.
\\The most important point in the design of FPAD system is to provide enough live and fake data to ``train'' the classifier, since the problem is treated as a pattern recognition problem, where the main modules are the capture device, the features extractor, the classifier \cite{Marasco:2014:SAS:2658850.2617756}. But these modules are useless without a representative set of data to tune their parameters, even if the feature extraction step is by-passed using a convolutional neural network as classifier.  

Fabricating spoofs is made difficult because this task leaves a large set of freedom degrees to the designer, from the basic method to obtain the so-called ``mold'', which can be cooperative or non-cooperative, the kind of materials adopted and the way to put the obtained spoofs (``cast'') on the sensor's surface \cite{ReviewLivDet}. A noticeable degree of cleverness and attention is necessary to find the way to fabricate a ``good'' fake fingerprint.

In this LivDet edition, we focused on three main problems with the aim to make more realistic the final reported performance. All cases are related to the possible behavior of the attackers and the information extracted from the users' fingerprints: (1) two groups of people were involved for fabricating spoofs and acquiring the related images: one for the training set, one for the test set; (2) two groups of materials were also adopted for the training set and the test set - in other words, the test set is made up of never-seen-before materials; (3) a subset of users was shared in both training set and test set - this choice was motivated by results reported in \cite{GhianiClientSpecific1}, in order to explore if user-specific information can improve the system's performance. The above points were not treated in the previous editions of LivDet, whose history and results over the years are reported in \cite{ReviewLivDet}. Last but not least, in this competition edition, the increasing interest on the fingerprint presentation attacks detection kept growing the number of participants, which is the biggest one ever reached before.

With regard to item (1), the presence of different operators for creating the fake fingerprints present in the train set and test set allows to simulate a real application scenario in which the operator training the system is different from the attackers.
For the same reason different materials for train set and test set have been used, according to item (2). In a realistic scenario, the system is not expected to be trained on the same materials used for a presentation attack.

Another important difference with respect to the previous editions of the competition is the presence of some users in both the training and testing parts of the three datasets. As a matter of fact, in the other editions, train and test sets were totally separated since none of the users were present in both parts.
However, ref. \cite{GhianiClientSpecific1} showed by experiments that the presence of different acquisitions of the same fingerprint in both parts of the dataset led to much better classification results with respect to a system were the users in train and test were different. We called the former a ``User Specific'' system and the latter a ``General Purpose'' one. The user specific information is particularly relevant when designing a fingerprint presentation attacks detector which must be integrated on a fingerprint verification system. In this case, it is expected to have some knowledges of the users stored into the system, and this knowledge can be exploit to design a software module ``intrinsically robust'' to presentation attacks when authenticating subjects.

\section{Participants, data sets and protocols}
\label{sec:Protocol}


\subsection{Participants}
\vspace{-5pt}
The participants to the LivDet competition are universities and companies that have a biometric liveness detection solution. After the registration phase, each competitor must signed a database release agreement detailing the proper usage of data made available through the competition. 

In Table \ref{tab:partecipantsLivDet} the participants names and the correspondent algorithms names are presented as they are used in this paper. In this competition many institutions made two software solutions. Moreover some competitors made GPU version in order to speed up the elaboration time of the algorithm. However this was not a constrain in the competition.
\begin{table}[h!]
\centering
\begin{tabular}{|c|c|}
\hline
\textbf{Participants} & \textbf{Algorithm names} \\ \hline
Suprema ID Inc.	&	SSLFD	\\ \hline
Hangzhou Jinglianwen Technology Co.,Ltd 	&	JLW\_A	\\ \hline
Hangzhou Jinglianwen Technology Co.,Ltd 	&	JLW\_B	\\ \hline
OKI Brasil	&	OKIBrB20	\\ \hline
OKI Brasil	&	OKIBrB30	\\ \hline
Zhejiang University of Technology	&	ZYL\_1	\\ \hline
Zhejiang University of Technology	&	ZYL\_2	\\ \hline
Anonymous 0	&	SNOTA2017\_1\\ \hline
Anonymous 0	&	SNOTA2017\_2\\ \hline
ModuLAB.	&	ModuLAB	\\ \hline
Chosun University	&	ganfp	\\ \hline
Anonymous 1	&	PB\_LivDet\_1	\\ \hline
Anonymous 1	&	PB\_LivDet\_2	\\ \hline
KAIST	&	hanulj	\\ \hline
Anonymous 2	&	SpoofWit	\\ \hline
University of Naples Federico II & LCPD \\ \hline
CENATAV & PADfV \\ \hline
\end{tabular}
\caption{ Name of the participants and the submitted algorithms.}
\label{tab:partecipantsLivDet}
\end{table}

\subsection{Data Sets}
\vspace{-5pt}
\begin{table*}[!t]
\centering
\begin{tabular}[t]{ | l | l | c | c | c | c |}
\hline
\textbf{Scanner} & \textbf{Model} & \textbf{Resolution [dpi]} & \textbf{Image Size [px]} & \textbf{Format} &\textbf{Type} \\ \hline
Green Bit & DactyScan84C & 500 & 500x500 & PNG & Optical \\ \hline
Orcanthus & Certis2 Image & 500 & 300x$n$  & PNG & Thermal swipe \\ \hline
Digital Persona & U.are.U 5160 & 500 &252x324 & PNG & Optical \\ \hline
\end{tabular}
\caption{Device characteristics for LivDet2017 datasets.}
\label{tab:sensors}
\end{table*}

\begin{table*}[t]
\centering
\resizebox{\textwidth}{!}{%
\begin{tabular}{|c|c|c|c|c|c|c|c|c|c|c|c|c|}
\hline
                 & \multicolumn{4}{c|}{\textbf{Train}}      & \multicolumn{4}{c|}{\textbf{Test}}\\ \hline
\textbf{Dataset} & Live & Wood Glue & Ecoflex & Body Double & Live & Gelatine & Latex & Liquid Ecoflex\\ \hline
Green Bit        &1000&400&400&400&1700&680&680&680\\ \hline
Orcanthus        &1000&400&400&400&1700&680&658&680\\ \hline
Digital Persona  &999&400&400&399&1700&679&670&679\\ \hline
\end{tabular}%
}
\caption{Number of samples for each scanner and each part of the dataset.}
\label{tab:datasetComposition}
\end{table*}

In order to build the data sets, we used three different scanners: Green Bit DactyScan84C, Orcanthus Certis2 Image and Digital Persona U.are.U 5160. The scanner characteristics are reported in Table \ref{tab:sensors}. Each device is more suitable for a different task. In particular, DactyScan84C is used for Italian border controls and issuance of italian electronic documents. The U.are.U 5160 scanner is inserted in the competition due to the possibility to use it with a mobile device. As a matter of fact we collected the fingerprint image using the Nexus7 tablet. Certis2 Image can be embedded in a PC. For Certis2 Image the images height is not constant but depends on the finger swipe way.
The use of these three scanners allows us to simulate different application contexts.

For each of these scanner we collected few less than 6000 images.  Live images came from multiple acquisitions of all fingers of different subjects. The LivDet 2017 fake images were collected using the cooperative method. 
Each dataset consists of two parts, the first is the train set, and the second is the test set. The train set is used to configure the algorithm while the algorithms performance are evaluated using the test set.
Moreover, the materials used in the training set are different with respect to the test set as we can see in Table \ref{tab:datasetComposition}. In this LivDet edition we also added a new peculiarity: the train set fake samples are built by an operator, and the fake samples in the test set are built by two other persons. In this way we simulate a real scenario, where the person ability that create the train set is different from that of possible attackers. In our opinion the manual ability is very important in order to attack an anti-spoofing system.

\subsection{Algorithms Submission}
\vspace{-5pt}
The  algorithms submission process for LivDet 2017 is same of previous editions. Each submitted algorithm is a console application with the following list of parameters:

\textit{[nameAlgorithm].exe [ndataset] [inputfile] [outputfile]}
\\The first parameter is the identification number of the dataset to analyze, the second is a list of absolute paths of each image to process, and the last is the path of an output file where the algorithm saves the classification result of every image. The scores are in the range from 0 to 100 where 0 means that the image is fake and 100 is the maximum degree of liveness. The selected classification threshold in order to measure the performance is 50. In case the algorithm has not been able to process the image, the corresponding output will be -1000.

Participants trained their algorithms on the training set. We tested them on the test sets and reported the following statistic measurements:

\begin{figure*}[h]
     \centering
     	\subfloat[]{\includegraphics[height=0.2\textheight]{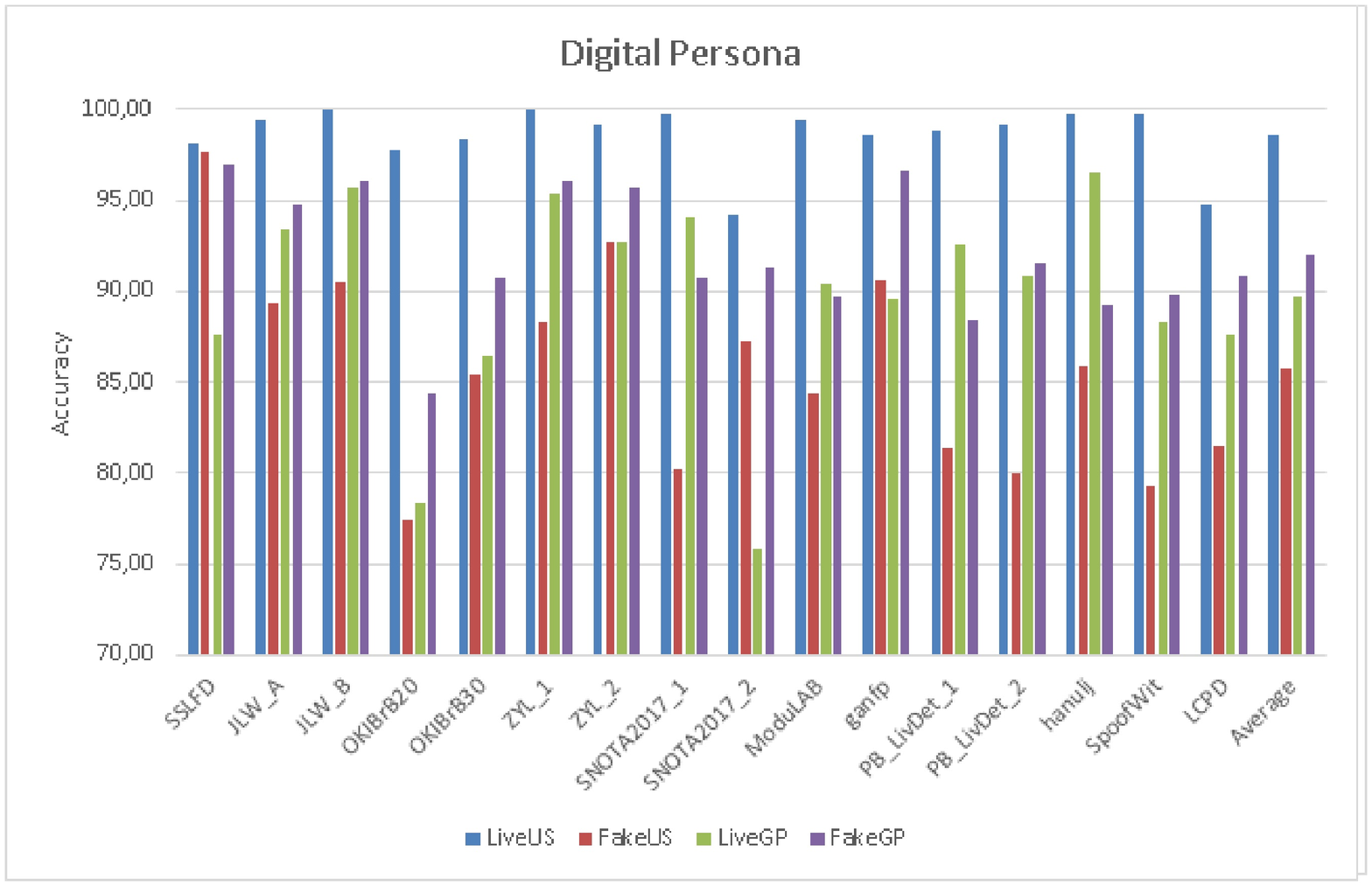}\label{<figure1>}}
        \vspace{-5pt}
     	\subfloat[]{\includegraphics[height=0.2\textheight]{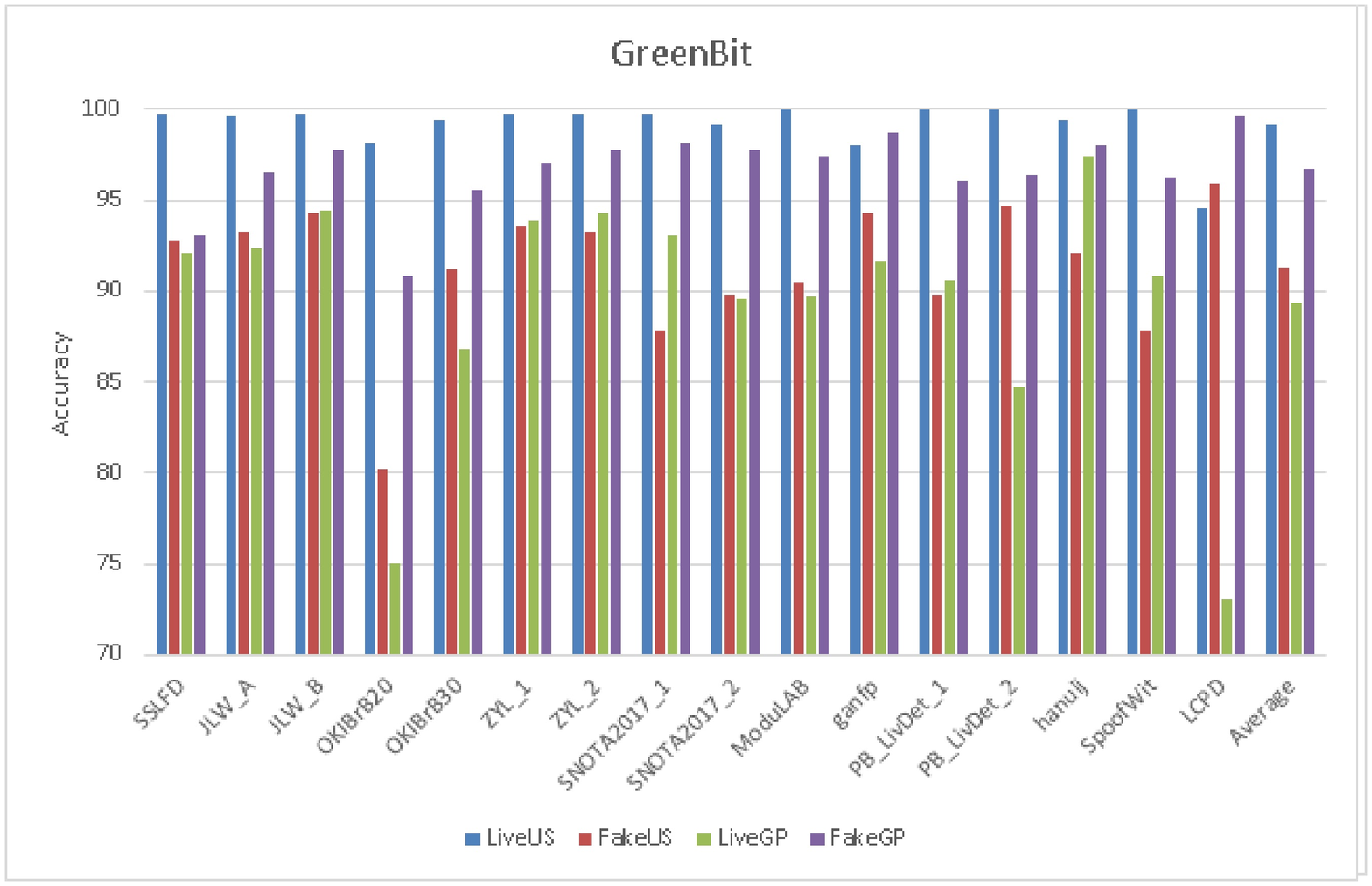}\label{<figure2>}}\\
        \vspace{-5pt}
     	\subfloat[]{\includegraphics[height=0.2\textheight]{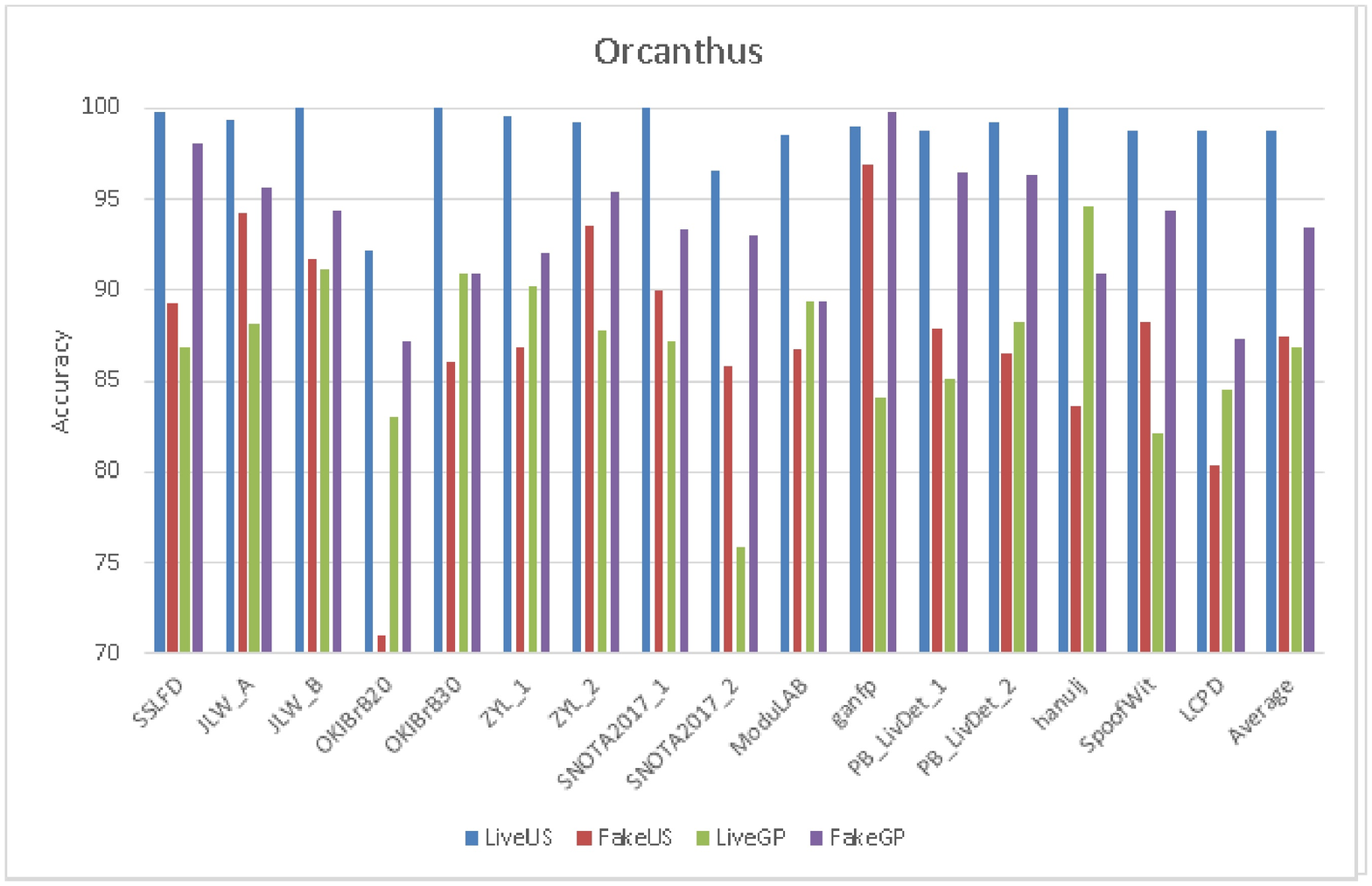}\label{<figure3>}}               
     	\caption{Algorithms accuracy rate of the   of lives and fakes in the User Specific (US) and General Purpose (GP) cases among using three LivDet competition scanners.}     
        \label{dpCS} 
\end{figure*}
The parameters adopted for the performance evaluation are the followings:
\begin{itemize}
\item \textit{Frej\_n}: Rate of failure to enroll, that is, the impossibility to process the fingerprint image and obtain the related features set.
\item \textit{Fcorrlive\_n}:  Rate of correctly classified live fingerprints.
\item \textit{Fcorrfake\_n}: Rate of correctly classified fake fingerprints.
\item \textit{Acc. Gelatine}: Rate of the correctly detected attacks with gelatine-based spoofs.
\item \textit{Acc. Latex}: Rate of the correctly detected attacks with latex-based spoofs.
\item \textit{Acc. Liquid Ecoflex}: Rate of the correctly detected attacks with liquid ecoflex-based spoofs.

\end{itemize}

\section{Discussions on the reported results}
\vspace{-5pt}
\label{sec:Result}
\begin{table}[!t]
\centering
\begin{tabular}[t]{ | l || c | c | c | c | r |}
\hline
\textbf{Algorithm}	& \textbf{1} & \textbf{2}	& \textbf{3}	& \textbf{Overall}\\ \hline
SSLFD	&93.58&94.33&93.14&93.68\\ \hline
JLW\_A	&95.08&94.09&93.52&94.23\\ \hline
JLW\_B	&96.44&95.59&93.71&95.25\\ \hline
OKIBrB20	&84.97&83.31&84.00&84.09\\ \hline
OKIBrB30	&92.49&89.33&90.64&90.82\\ \hline
ZYL\_1	&95.91&95.13&91.66&94.23\\ \hline
ZYL\_2	&96.26&94.73&93.17&94.72\\ \hline
SNOTA2017\_1	&95.03&91.26&91.58&92.62\\ \hline
SNOTA2017\_2	&94.04&86.72&86.74&89.17\\ \hline
ModuLAB	&94.25&90.40&90.21&91.62\\ \hline
ganfp	&95.67&93.66&94.16&94.50\\ \hline
PB\_LivDet\_1	&93.85&89.97&91.85&91.89\\ \hline
PB\_LivDet\_2	&92.86&90.43&92.60&91.96\\ \hline
hanulj	&97.06&92.34&92.04&93.81\\ \hline
SpoofWit	&93.66&88.82&89.97&90.82\\ \hline
LCPD &89.87&88.84&86.87&88.52\\ \hline
PDfV&92.86&93.31&N.A.&N.A.\\ \hline
\end{tabular}
\caption{Accuracy of the algorithms on the test sets [\%]. \textbf{1} = Green Bit, \textbf{2} = Digital Persona, \textbf{3} = Orcanthus.}
\label{tab:overall}
\end{table}

\begin{table*}[!tb]
\centering
\resizebox{\textwidth}{!}{%
\begin{tabular}[t]{ |c| l || c | c | c | c | c | c | r |}\hline
&\textbf{Algorithm}	& \textbf{Frej} & \textbf{Fcorrlive}	& \textbf{Fcorrfake}	& \textbf{Acc.}	& \textbf{Acc.} & \textbf{Acc. Liquid}& \textbf{Accuracy}\\
& & \textbf{[\%]}	&	\textbf{[\%]}	& \textbf{[\%]}	&	\textbf{Gelatine [\%]}	&	\textbf{Latex [\%]}	&		\textbf{Ecoflex [\%]} & \textbf{[\%]}\\ \hline 
\multirow{12}*{\begin{sideways}\textbf{Green Bit}\end{sideways}}
&SSLFD	&0.00&94.71&92.65&	 85.29	&	 95.29	&	 98.38 	&
93.58\\ 
&JLW\_A	&0.00&94.47&95.59&	 91.47	&	 96.03	&	 99.26 	&
95.08\\ 
&JLW\_B	&0.00&96.06&96.76&	 91.62	&	 98.68	&	100.00 	&
96.44\\ 
&OKIBrB20&0.00&81.88&87.55&	 77.94	&	 85.88	&	 99.26 	&
84.97\\ 
&OKIBrB30&0.00&91.23&93.53&	 90.15	&	 94.12	&	 98.53 	&
92.49\\ 
&ZYL\_1&0.00&95.65&96.13&	 90.88	&	 97.65	&	 99.85 	&
95.91\\ 
&ZYL\_2&0.00&95.94&96.52&	 92.94	&	 96.91	&	 99.71 	&
96.26\\ 
&SNOTA2017\_1&0.00&95.23&94.85&	 93.53	&	 92.35	&	 99.41 	&
95.03\\ 
&SNOTA2017\_2&0.00&93.76&94.26&	 91.62	&	 95.44	&	 99.41 	&
94.04\\ 
&ModuLAB&0.00&92.94&95.34&	 92.35	&	 96.03	&	 97.94 	&
94.25\\ 
&ganfp&0.00&93.53&97.45&	 96.47	&	 95.88	&	100.00 	&
95.67\\ 
&PB\_LivDet\_1&0.00&93.41&94.21&	 88.09	&	 94.85	&	 99.71 	&
93.85\\ 
&PB\_LivDet\_2&0.00&89.23&95.88&	 91.76	&	 96.18	&	 99.85 	&
92.86\\ 
&hanulj&0.00&98.06&96.22&	 93.53	&	 96.62	&	 98.68 	&
97.06\\ 
&SpoofWit&0.00&93.70&93.63&	 90.29	&	 91.62	&	 99.41 	&
93.66\\ 
&LCPD&0.00&79.41&98.58&	 98.82	&	 96.91	&	100.00 	&
89.87\\ 
&PDfV&0.00&92.41&93.23&	 86.62	&	 94.26	&	 98.82 	&
92.86\\ \hline 
\multirow{12}*{\begin{sideways}\textbf{Digital Persona}\end{sideways}}
&SSLFD	&0.00&91.19&96.94&	 97.94	&	 93.67	&	 99.25 	&
94.33\\ 
&JLW\_A	&0.00&95.15&93.19&	 87.63	&	 92.05	&	100.00 	&
94.09\\ 
&JLW\_B	&0.00&96.99&94.43&	 87.33	&	 96.02	&	100.00 	&
95.59\\ 
&OKIBrB20&0.00&84.69&82.15&	 70.99	&	 81.00	&	 94.63 	&
83.31\\ 
&OKIBrB30&0.00&90.66&88.21&	 77.17	&	 88.66	&	 98.96 	&
89.33\\ 
&ZYL\_1&0.00&96.75&93.79&	 85.86	&	 95.58	&	100.00 	&
95.13\\ 
&ZYL\_2&0.00&94.62&94.82&	 91.75	&	 92.78	&	100.00 	&
94.73\\ 
&SNOTA2017\_1&0.00&95.80&87.48&	 82.03	&	 81.00	&	 99.55 	&
91.26\\ 
&SNOTA2017\_2&0.00&85.11&88.07&	 88.66	&	 77.91	&	 97.76 	&
86.72\\ 
&ModuLAB&0.00&93.26&88.02&	 85.57	&	 80.56	&	 98.06 	&
90.40\\ 
&ganfp&0.00&92.20&94.87&	 90.28	&	 94.40	&	100.00 	&
93.66\\ 
&PB\_LivDet\_1&0.00&94.39&86.29&	 69.96	&	 89.99	&	 99.10 	&
89.97\\ 
&PB\_LivDet\_2&0.00&93.32&88.02&	 79.68	&	 85.71	&	 98.81 	&
90.43\\ 
&hanulj&0.00&97.52&88.02&	 86.60	&	 81.30	&	 96.27 	&
92.34\\ 
&SpoofWit&0.00&92.20&86.00&	 77.03	&	 82.03	&	 99.10 	&
88.82\\ 
&LCPD&0.00&89.72&88.12&	 85.27	&	 80.41	&	 98.81 	&
88.84\\ 
&PDfV&0.00&93.56&93.10&	 91.02	&	 89.25	&	 99.10 	&
93.31\\ \hline 
\multirow{12}*{\begin{sideways}\textbf{Orcanthus}\end{sideways}}
&SSLFD	&0.00&91.12&94.85&	 91.76	&	 96.35	&	 96.47 	&
93.14\\ 
&JLW\_A	&0.00&91.47&95.24&	 88.97	&	 97.26	&	 99.56 	&
93.52\\ 
&JLW\_B	&0.00&93.76&93.66&	 83.24	&	 98.48	&	 99.41 	&
93.71\\ 
&OKIBrB20&0.00&86.00&82.31&	 79.56	&	 74.77	&	 92.35 	&
84.00\\ 
&OKIBrB30&0.00&93.94&87.86&	 79.85	&	 90.73	&	 93.09 	&
90.64\\ 
&ZYL\_1&0.00&92.94&90.58&	 76.47	&	 96.66	&	 98.82 	&
91.66\\ 
&ZYL\_2&0.00&91.12&94.90&	 88.24	&	 97.11	&	 99.41 	&
93.17\\ 
&SNOTA2017\_1&0.00&91.12&91.97&	 81.32	&	 95.90	&	 98.82 	&
91.58\\ 
&SNOTA2017\_2&0.03&84.88&88.31&	 74.56	&	 92.55	&	 97.94 	&
86.74\\ 
&ModuLAB&0.00&92.24&88.50&	 75.15	&	 91.95	&	 98.53 	&
90.21\\ 
&ganfp&0.00&88.47&98.96&	 97.65	&	 99.24	&	100.00 	&
94.16\\ 
&PB\_LivDet\_1&0.00&89.29&94.00&	 89.71	&	 95.44	&	 96.91 	&
91.85\\ 
&PB\_LivDet\_2&0.00&91.53&93.51&	 89.56	&	 94.07	&	 96.91 	&
92.60\\ 
&hanulj&0.00&96.35&88.40&	 76.32	&	 93.47	&	 95.59 	&
92.04\\ 
&SpoofWit&0.00&87.65&91.92&	 84.71	&	 92.25	&	 98.82 	&
89.97\\ 
&LCPD&0.00&88.71&85.33&	 69.85	&	 88.75	&	 97.50 	&
86.87\\ 
&PDfV&100.00&N.A.&N.A.&N.A.&N.A.&N.A.&N.A\\ \hline 

\end{tabular}
}
\caption{Performance of all algorithms on all datasets.}
\label{tab:performance}
\end{table*}

The correct classification rates of each algorithm on the LivDet test sets, including the overall average accuracy, are summarized in Table \ref{tab:overall}.
It is worth noting that all algorithms achieved over 80\% accuracy in all test sets. 
We can appreciate that the accuracy of the GreenBit dataset is higher than that achieved in the other ones. This is very important because the security of national borders is a crucial issue. A possible explanation is due to the image size given by the scanner: in Table \ref{tab:sensors}, we can see that GreenBit image size is almost twice the one of the other scanners. Thus, it may be hypothesized that the feature extractor, or the convolutional layers of the classifiers, where employed, can analyze more live/fake discriminating details. On the other hand, we may expect that when the sensor must be integrated on the personal verification device on the basis of the application requirements, a solid state sensor is still preferable, at the expense of less fingerprint area to be captured. Therefore, according to the reported results, a small drop of performance may occur.  \\
Unfortunately, the PDAfV algorithm was not able to process the Orcanthus images and for this reason it did not compete for the final win. This also explained that the value of $100.00\%$ is reported on the last row of Table \ref{tab:performance}.
\\Table \ref{tab:performance} reports the details of the performance competition. It is divided in three subtables, for each LivDet data set. The parameters in columns have been described in the previous Section.

All algorithms classified all the images, except SNOTA2017\_1 which has not been able to analyze one image from the Orcanthus dataset. For the Digital Persona test set, the majority of the algorithms has $Fcorrlive > Fcorrfake$, whilst the other test sets has the $Fcorrlive < Fcorrfake$. This may be not considered as a ``problem'' because it is probable that the threshold we adopted for the final decision should be increased to obtain the same rate of correctly detected spoofs, at the expense of the $Fcoorlive$ reduction. 
This points out that all software-based approaches should be designed by considering the overall application: since a false positive detection, that is, the misclassification of a live fingerprint as a spoof, significantly impacts on the overall performance of the fingerprint verification system, a way to set an appropriate trade-off operational points for FPAD and verification modules should be investigated. 


In Table \ref{tab:performance} it is also worth noting the different accuracy rates related to the three never-seen-before materials. While all the algorithms easily recognized the majority of Liquid Ecoflex fakes, the other two materials resulted to be much tougher to classify. Latex and, even more, Gelatine strongly contributed to the accuracy reduction. This drop of performances is particularly emphasized in the case of the Digital Persona and Orcanthus devices.

Finally, the results of the ``user specific'' vs. ``generic users'' experiments are reported in Figures \ref{dpCS}.a, \ref{dpCS}.b and \ref{dpCS}.c. In the three set of histograms, the accuracy rate of lives and fakes are reported.
In each figure the average of the four accuracy values is also presented. As expected, the presence in the training of different acquisitions of the same fingerprint that has to be classified resulted in a better accuracy rate. This is evident in the case of the live samples. Regarding the fake samples the accuracy decrease is due to the different expertise of the persons in charge of their fabrication. It is worth remarking that this is not a ``training trick'', because we are referring to the application of fingerprint verification when the user population is stored in the system's database; thus the exploitation of information coming from such users can be done.
For lack of space we were unable to present all the experimental results that will be inserted in a future complete report.

\section{Conclusion}
    \vspace{-5pt}
\label{sec:Conclusion}
The fifth edition of LivDet 2017 proved to be a competition of increasing success. This edition focused on pointing out the relationships between obtained performance and characteristics of attackers and users. The number of software solution is the biggest ever reached, and the performance showed that this arms-race problem still needs a lot of work, since the best performance achieved is around $95\%$ of correct classification rate. We hope that findings summarized in this report can be of some helps for academies and companies working on presentation attacks detection. However, a deep analysis of these LivDet results is needed in order to better understand merits and limitations of current state-of-the-art approaches. This will be done in a future work.

\section{Acknowledgements}
 \vspace{-5pt}
This edition has been possible thanks to the sponsorship of the GreenBit company. 

We want to thank Stephanie Schuckers, who suggested us the initial insights to the fabrication of fake fingerprints in 2006 and cooperated to the organization of the first competition (2009) up to the following ones (2011, 2013, 2015). Her help was significant for allowing us to autonomously gain, over time, a deep knowledge on the problem of fingerprint presentation attacks detection and contributed to the state-of-the-art.


\begin{thebibliography}{1}

\bibitem{GhianiClientSpecific1}
L.~Ghiani, G.~L. Marcialis, F.~Roli, and P.~Tuveri.
\newblock User-specific effects in fingerprint presentation attacks detection:
  Insights for future research.
\newblock In {\em 2016 International Conference on Biometrics (ICB)}, pages
  1--6, June 2016.

\bibitem{ReviewLivDet}
Luca Ghiani, David~A. Yambay, Valerio Mura, Gian~Luca Marcialis, Fabio Roli,
  and Stephanie~A. Schuckers.
\newblock Review of the fingerprint liveness detection (livdet) competition
  series: 2009 to 2015.
\newblock {\em Image and Vision Computing}, 58(Supplement C):110 -- 128, 2017.

\bibitem{kalloPatent}
P.~Kallo, I.~Kiss, A.~Podmaniczky, and J.T. Losi.
\newblock Detector for recognizing the living character of a finger in a
  fingerprint recognizing apparatus, January~16 2001.
\newblock US Patent 6,175,641.

\bibitem{Marasco:2014:SAS:2658850.2617756}
Emanuela Marasco and Arun Ross.
\newblock A survey on antispoofing schemes for fingerprint recognition systems.
\newblock {\em ACM Comput. Surv.}, 47(2):28:1--28:36, November 2014.

\bibitem{MatsumotoMYH02}
Tsutomu Matsumoto, Hiroyuki Matsumoto, Koji Yamada, and Satoshi Hoshino.
\newblock Impact of artificial "gummy" fingers on fingerprint systems.
\newblock {\em Datenschutz und Datensicherheit}, 26(8), 2002.

\bibitem{Schuckers02spoof}
Stephanie A.~C. Schuckers, Stephanie A.~C. Schuckers, Ph. D, and Ph. D.
\newblock Spoofing and anti-spoofing measures.
\newblock {\em Information Security Technical Report}, 7:56--62, 2002.

\bibitem{vanderPutte2000}
Ton van~der Putte and Jeroen Keuning.
\newblock {\em Biometrical Fingerprint Recognition: Don't get your Fingers
  Burned}, pages 289--303.
\newblock Springer US, Boston, MA, 2000.

\bibitem{Scanners}
David Willis and Mike Lee.
\newblock Six biometric devices point the finger at security.
\newblock {\em Netw. Comput.}, 9(10):84--96, June 1998.

\end{thebibliography}

\end{document}